\documentclass[runningheads]{llncs}
\usepackage{graphicx}
\usepackage{caption}
\usepackage{subcaption}
\usepackage{amsmath}
\usepackage{authblk}
\usepackage{xcolor}
\usepackage{cite}
\usepackage{url}
\usepackage{amsmath}
\usepackage{soul}
\usepackage[export]{adjustbox}

\begin{document}
\title{3D Semantic Mapping from Arthroscopy using Out-of-distribution Pose and Depth and In-distribution Segmentation Training}
\author{Yaqub Jonmohamadi\inst{1} \and
Shahnewaz Ali\inst{1} \and
Fengbei Liu\inst{2} \and
Jonathan Roberts\inst{1} \and
Ross Crawford\inst{1} \and
Gustavo Carneiro\inst{2} \and
Ajay K. Pandey \inst{1}}
\authorrunning{Jonmohamadi et al.}
%

\institute{School of Electrical Engineering and Robotics, Science and Engineering Faculty, Queensland University of Technology, \quad
\email{y.jonmo@gmail.com} \and
Australian Institute for Machine Learning, School of Computer Science, University of Adelaide \quad \email{gustavo.carneiro@adelaide.edu.au}}


\maketitle              
\begin{abstract}

Minimally invasive surgery (MIS) has many documented advantages, but the surgeon's limited visual contact with the scene can be problematic. Hence, systems that can help surgeons navigate, such as a method that can produce a 3D semantic map, can compensate for the limitation above. In theory, we can borrow 3D semantic mapping techniques developed for robotics, but this requires finding solutions to the following challenges in MIS: 
1) semantic segmentation, 2) depth estimation, and 3) pose estimation. In this paper, we propose the first 3D semantic mapping system from knee arthroscopy that solves the three challenges above.
Using out-of-distribution non-human datasets, where pose could be labeled, we jointly train depth+pose estimators using self-supervised and supervised losses. Using an in-distribution human knee dataset, we train a fully-supervised semantic segmentation system to label arthroscopic image pixels into femur, ACL, and meniscus. Taking testing images from human knees, we combine the results from these two systems to automatically create 3D semantic maps of the human knee. The result of this work opens the pathway to the generation of intra-operative 3D semantic mapping, registration with pre-operative data, and robotic-assisted arthroscopy.

\keywords{3D Semantic Mapping, Endoscopy, Deep Learning.}
\end{abstract}

\section{Introduction}
\label{sec:introduction}

Minimally invasive surgery (MIS) is a surgical procedure where the operation is conducted via a few incision holes. It is favorable over open surgery due to its clinical benefits such as small scars, lower chances of bleeding and infection, and shorter recovery time. 
However, MIS forces the surgeon to lose direct eye contact with the scene and, consequently, to rely on endoscopic video for the whole surgery. 
The limited field of view (FoV) and 2D nature of endoscopic images are challenges that surgeons face which quite often result in surgeons failing to identify the tissue structures and recourse to visual surveying by moving the camera around. In arthroscopy, this phenomenon happens repeatedly during surgery, which could prolong the operation time and lead to unintentional damage to critical tissue. According to a survey on knee arthroscopy \cite{jaiprakash_orthopaedic_2017}, about 50\% of the surgeons admitted to damage a knee once every 10 operations. 

Computer vision could assist surgeons by augmenting the reality produced by the endoscopic image with the creation of a 3D semantic map of the scene,  color-coded to represent the different anatomical structures and surgical tools. 
To produce a 3D semantic map in arthroscopy, we need to solve the following challenges: semantic segmentation, depth estimation, and pose estimation. 
In other types of MIS, like laparascopy and sinus endoscopy, techniques such as simultaneous localization and mapping (SLAM) \cite{grasa_visual_2013} and structure from motion (SfM) \cite{leonard_evaluation_2018} have been applied successfully.
However, such techniques will fail in arthroscopy, due to poor texture, lack of photometric constancy across the frames, and assumptions regarding the camera motion.

 In this paper, we introduce the first method to produce 3D semantic maps from arthroscopy.
 To achieve this goal, we create two datasets: 1) out-of-distribution (OOD) datasets containing non-human knees that have camera pose annotation, and 2) an in-distribution (ID) dataset containing human knees that have semantic segmentation annotation of femur, Articular Cartilage Ligament (ACL), and meniscus.
 We train a system with the OOD datasets to estimate depth+pose using self-supervised view synthesis loss + supervised pose loss.
 We also train a method to produce semantic segmentation using the ID dataset in~\cite{jonmohamadi_automatic_2020}.
 We then combine the pose, depth, and semantic segmentation of both systems and use the method in~\cite{zach_globally_2007} to produce 3D semantic maps of testing images from the ID dataset.
 Quantitative results of the pose estimation and qualitative visual results from the 3D semantic maps suggest that our approach can be reliably used for mapping human knees, even though part of the training was based on OOD training sets. To the best of our knowledge, this is the first method that can estimate the depth and pose from arthroscopy and the first to create  3D semantic maps in clinical endoscopy.

\section{Related Work}
\vspace{-1mm}
Deep learning has shown impressive results in complex computer vision tasks such as segmentation, depth perception, and pose estimation~\cite{vijayanarasimhan_sfm-net_2017,godard_digging_2019,zhou_unsupervised_2017}. 
These approaches work well on feature rich datasets like road scenes but perform poorly for environments such as medical endoscopy as shown in~\cite{sharan_domain_2020}.
This is because of poor texture information and the lack of photometric constancy between frames in endoscopy due to the joint motion between the camera and light source~\cite{liu_dense_2019}.
Recently, depth and pose estimation methods above have been adapted for colonoscopy \cite{chen_slam_2019,bae_deep_2020,ma_real-time_2019} and sinus endoscopy \cite{liu_dense_2019,liu_self-supervised_2018,liu_reconstructing_2020}. Compared to the original work on self-supervised estimation of depth and pose shown in \cite{vijayanarasimhan_sfm-net_2017,godard_digging_2019,zhou_unsupervised_2017}, a key aspect to these proposed methods is the incorporation of supervision for depth+pose estimation. 
For example, \cite{bae_deep_2020,ma_real-time_2019,liu_dense_2019,liu_self-supervised_2018,liu_reconstructing_2020} used structure from motion (SfM)~\cite{ullman_interpretation_1979}  to create sparse depth frames from the training images and used them for supervision of the depth+pose training.
Arthroscopy images have little texture information due to the smooth bone surfaces. 
Furthermore, the problem of over and under illumination in arthroscopy is a frequent occurrence that will impact the approaches above~\cite{ali_supervised_2020}. 
As a result, feature tracking based techniques such as SfM, cannot create reliable feature maps in arthroscopy as has been shown in~\cite{marmol_dense-arthroslam_2019}. 




Hence, we advocate the use of pose annotation acquired from images from non-human environments to supervise the training of depth+pose using a self-supervised+supervised loss function. We also trained a novel supervised model for semantic segmentation with the method in~\cite{Shahnewaz_arthos_2021} that extends the semantic segmentation in~\cite{jonmohamadi_automatic_2020} based on the use of multi-spectral frame reconstruction~\cite{Otsu_2018}. 
By considering that the biological compositions of each tissue type namely bone, ACL, and meniscus are intrinsically different, the RGB arthroscopic frames are transformed into 36 spectral bands and then the spatial features of anatomical structures are used at wavelengths from 380-740 nm with 10 nm of intervals as a preprocessing step. 
A segmentation network extracts spatial characteristics at these 36 spectral bands and subsequently learns the location along with its label. 

\section{Methods}
The aim of depth + pose network, for a given source image at time $t$, $\mathbf{I}_t$,  and source frames, $\mathbf{I}_S$, is to estimate the pixel level depth $\hat{\mathbf{D}}_t =f_D (\mathbf{I}_t)$ and the ego motion $\mathbf{X}^l_{t\rightarrow{t+\Delta t}} = [{Rot} ~ {Trl}]$, where $Rot=[\alpha, \beta, \gamma]$ and $Trl=[x, y, z]$ refer to the 6 degree of freedom, rotation and translation, in the Euler coordinates 
We achieve this by training the depth+pose network on the self-supervised plus supervised objectives. In our case, with a stereo endoscope, the source images $\mathbf{I}_S$ are the left image at time $t+1:\mathbf{I}_{t+1}^l$ and the right image at time $t:\mathbf{I}_t^r$, while the target image is the left image at time $t:\mathbf{I}_t^l$.

\subsubsection{Self-supervised objective}
 minimizes a photometric reprojection error between the synthesized target image, $\hat{\mathbf{I}}_t$ and the target image, $\mathbf{I}_t$ as shown in \cite{godard_unsupervised_2017,zhou_unsupervised_2017} and edge-aware smoothing term as shown in \cite{yang_lego_2018,godard_digging_2019}:
\begin{equation}
L_{self}(\mathbf{I}_t, \hat{\mathbf{I}}_t) = L_{phot}(\mathbf{I}_t, \mathbf{I}_S)\odot \mathbf{M}_{phot} + \lambda _{smoo}L_{smoo}(\hat{\mathbf{D}}) 
\end{equation}
where $L_{phot}(\mathbf{I}_t, \hat{\mathbf{I}}_t)$ is the pixel level photometric reprojection error shown in \cite{godard_unsupervised_2017} and consist of structural similarity (SSIM) term \cite{wang_image_2004} and a $L1$ loss: 
\begin{equation}
L_{phot}(\mathbf{I}_t, \hat{\mathbf{I}}_t) = \alpha~\frac{1-SSIM(\mathbf{I}_t, \hat{\mathbf{I}}_t)}{2} + (1-\alpha)\|\mathbf{I}_t-\hat{\mathbf{I}}_t\|,
\end{equation}

where $\alpha=0.85$. Similar to \cite{godard_digging_2019}, the minimum reprojection error is used to minimize the effect of the pixels which are not visible in some of the source images compared with the target image due to ego motion or occlusion: 
\begin{equation}
L_{phot}(\mathbf{I}_t, \mathbf{I}_s) = \min_{\mathbf{I}_S} L_{phot}(\mathbf{I}_t, \hat{\mathbf{I}}_t). 
\end{equation}
The minimum reprojection loss is particularly helpful in reducing the edge artifacts of the depth.   
The auto masking term $\mathbf{M}_{phot}$ is a binary mask to reject the pixels with no change in appearance between frames such as static scenes and the moving objects at the same velocity and orientation of the camera \cite{godard_digging_2019}:
\begin{equation}
\mathbf{M}_{phot} = \min_{\mathbf{I}_S} L_{phot}(\mathbf{I}_t, \mathbf{I}_s) > \min_{\mathbf{I}_S} L_{phot}(\mathbf{I}_t, \hat{\mathbf{I}}_t).
\end{equation}

Similar to \cite{godard_unsupervised_2017} the weighted edge-aware term $\lambda_{smoo} L_{smoo}(\hat{\mathbf{D}})$ is used to regularize the depth on low texture areas:
\begin{equation}
L_{smoo}(\hat{\mathbf{D}}) = |\delta_x \hat{\mathbf{D}}_t |e^{-|\delta_x \mathbf{I}_t |} + |\delta_y \hat{\mathbf{D}}_t |e^{-|\delta_y \mathbf{I}_t |} |,
\end{equation}
where $\delta$ refers to the gradient function is 1e-3.

\subsubsection{Supervised objective} 
is to minimize the error between the estimated ego motion by the pose network $\hat{\mathbf{X}}_{t\rightarrow{t+1}} = [\hat{Rot} ~ \hat{Trl}]$ and the groundtruth relative camera pose  ${\mathbf{X}}_{t\rightarrow{t+1}} = [{Rot} ~ {Trl}]$:
\begin{equation}
\begin{aligned}
L_{pose}(\mathbf{X}_{t\rightarrow t+1} , \hat{\mathbf{X}}_{t\rightarrow t+1}) = L_{Trl} + L_{Ang}, \quad and \\
L_{Trl} = L_{Trl}' + L_{Trl}'', ~~ L_{Trl}' = || Trl’ - \hat{Trl}’||,~  
L_{Trl}'' = ||Trl - \hat{Trl}||, and \\
L_{Ang} = L_{Ang}' + L_{Ang}'', ~~ L_{Ang}' = || Ang’ - \hat{Ang}’||,~  
L_{Ang}'' = ||Ang - \hat{Ang}||.
\label{LossSuper}
\end{aligned}
\end{equation}
The term $L_{Trl}'$ is calculated on the normalized translations, i.e., $Trl’ = Trl/||Trl||$ and $\hat{Trl}’ = \hat{Trl}/||\hat{Trl}||$. Similarly, $Ang’ = Ang/||Ang||$ and $\hat{Ang}’ = \hat{Ang}/||\hat{Ang}||$. This is because the relative displacement from frame to frame could substantially change for the endoscopic sequences with some frames having more than 20 times change in translation or angle compared with other frames. Without the $L_{Trl}'$, the network performs poorly for the frames with small changes in motion. 

\begin{figure}
  \begin{minipage}[c]{0.62\textwidth}
    \includegraphics[width=\textwidth]{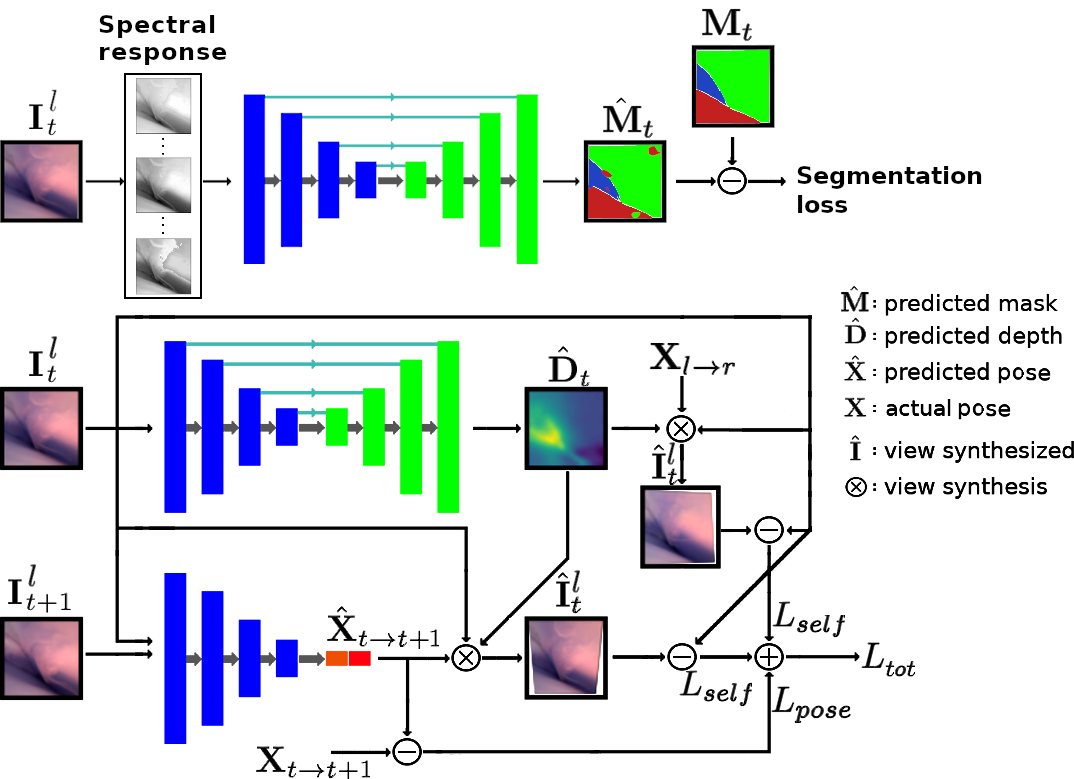}
  \end{minipage}\hfill
  \begin{minipage}[c]{0.35\textwidth}
    \caption{
       The training pipeline for the segmentation, depth and pose estimation. The upper network performs semantic segmentation in a supervised manner. The second and third networks are the depth+pose networks being trained jointly using the supervise+self-supervised approach.}
    \label{Pipeline}
  \end{minipage}
\end{figure}

The final loss equation for training the depth+pose network in this work is:
\begin{equation}
L_{tot}(\mathbf{I}_t, \hat{\mathbf{I}}_t) = L_{self}(\mathbf{I}_t, \hat{\mathbf{I}}_t) + L_{pose}(\mathbf{X}_{t\rightarrow t+1}, \hat{\mathbf{X}}_{t\rightarrow t+1})
\end{equation}
Since most of the variation in the camera pose is in at the x and y axes of the translation, the weighting of [0.5, 0.5, 1] was applied to the $L_{Trl}$.
Fig. \ref{Pipeline} shows the pipeline for training the depth+pose and segmentation networks.
\vspace{-2mm}
\vspace{-2mm}

The method in~\cite{zach_globally_2007} was used to fuse $n$ depth frames and create the 3D maps in chunks. 
Since in the actual arthroscopy, the endoscope movement is limited to areas surrounding the incision holes, the sequences used for fusion are typically 3 to 8 seconds long ~($n \in \{70,...,200\}$ frames), which correspond to sweep in translation to cover a certain part of the knee accessible via the incision hole. 



\section{Experimental setup}
\subsection{Depth+pose training}
 Training + validation data was recorded using the stereo endoscope with 384$\times$384 resolution, 1.52 mm baseline, 87.5$^{\circ}$ FoV, and 25 fps. Stereo images were rectified and downsampled to 256$\times$256. 
The groundtruth poses of the camera tip were recorded by attaching an NDI magnetic sensor. 

Liu et al.~\cite{liu_self-supervised_2020} showed that training a self-supervised depth network directly on the arthroscopy videos failed due to the poor texture and pretraining on texture rich frames is crucial. Therefore, for the training, a 3D printed model of the knee was placed in a water tank and recorded while the magnetic sensor was attached to the camera tip to provide the groundtruth camera poses. The images of the 3D printed knee provided rich texture and are ideal for the training phase. Another data was recorded from a sheep joint with the corresponding groundtruth camera poses. Similar to the human knee, the animal joint video frames suffer from the poor texture problem. During the training and validation, the images from the 3D printed knee and the animal joint were used at the same time. 

In total 28500 frames from the 3D printed knee and 10000 frames from the animal experiment were recorded, out of which 25000 from the 3D printed and 8800 from the animal experiment were used for training and the rest for validation. The Absolute Trajectory Error (ATE) was used to quantify the error between the network estimated and the groudtruth poses.

The testing data was obtained from a cadaveric experiment in which multiple sequences were recorded from a left knee. About 9 sequences were recorded with varying lengths, few seconds to a few minutes. This resulted in approximately 12000 images. No groundtruth camera pose was available for these sequences. 
 


The depth network architecture was similar to Disp-Net \cite{mayer_large_2016}, but also uses skip connection \cite{ronneberger_u-net_2015} from the encoder’s activation blocks, leading to higher resolution details \cite{godard_unsupervised_2017} with sigmoids at the output.
For the pose, only the encoder was used. The ResNet50 \cite{he_deep_2016} was used as the encoder for both depth and pose networks. The weights were pretrained on ImageNet \cite{russakovsky_imagenet_2015}.
The training augmentations was used with 50\% possibility of random brightness, contrast, saturation, and hue jitter with respective ranges of ±0.2, ±0.2, ±0.2, and ±0.1 \cite{godard_digging_2019}. The model was trained on 30 epochs, using Adam \cite{da_method_2014} optimizer, with a batch size of 18. Since most of the variation in the camera pose is in at the x and y axes of the translation, the weighting of [0.5, 0.5, 1] was applied to the translation loss.

\subsection{Segmentation training}
Data from four cadaveric experiments were used for training and testing. Data from the last experiment were used as the test data for the depth+pose networks~\cite{jonmohamadi_automatic_2020}. There were 2868 images from the first experiment and 1524 from the last experiment (two sequences among nine) that were used for training. The remaining images from the last experiment (3460 frames) were used for testing along with the other three cadaver experiments. We test on two sets: i) high quality of images, and ii) all remaining cadaver datasets, excluding saturated and bad frames. It has been confirmed that the accuracy of the proposed method can be improved if high quality imaging system and sufficient information about the irregular knee geometrical structures are provided. More details are available on \cite{Shahnewaz_arthos_2021}. The training data was augmented 6 times using shift and rotation (with angles 90, 180, 270), flip vertical and horizontal, and brightness changes.

The model in \cite{Shahnewaz_arthos_2021} is a U-Net with the contraction layer containing two successive convolution layers and a 3$\times$3 kernel. The spatial context map is downsampled by max pooling operation with pool size 2$\times$2. Padding 'same' is used to get the same resolution of input and output images. Kernel initializer is used to set initialize weights of the convolution layer during training. The dilation rate is set to 2 which provides a wider field of view so that it can avoid adjacent pixels having the same reflectance. The softmax is used at the final layer. Categorical cross-entropy loss function and Stochastic gradient descent optimizer are used for training. The Tensorflow \cite{paszke_automatic_2017} was used to implement all the models.

\section{Results}
\begin{figure}[t]

        \begin{subfigure}{1.15in}
                \includegraphics[trim={0mm 0mm 0mm 0mm},clip, scale=0.3]{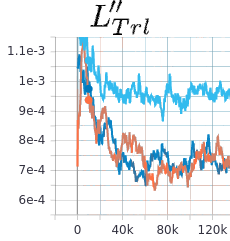} \hspace{0mm}
                \subcaption[]{Validation loss \\ translation}
        \end{subfigure} %
        \begin{subfigure}{1.15in}
                \includegraphics[trim={0mm 0mm 0mm 0mm},clip, scale=0.3]{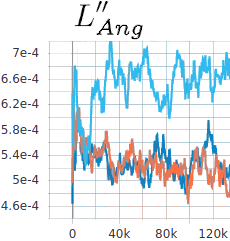} \hspace{0mm}
                \subcaption[]{Validation loss \\ angle}
        \end{subfigure} %
        \begin{subfigure}{1.3in}
            \includegraphics[trim={0mm 0mm 0mm 0mm},clip, scale=0.3]{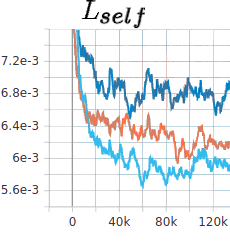} \hspace{0mm}
            \subcaption[]{Validation photometric reprojection loss}
        \end{subfigure} %
        \begin{subfigure}{0.905in}
            \includegraphics[trim={0mm 0mm 0mm 0mm},clip, scale=0.32]{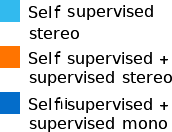} \hspace{0mm}
        \end{subfigure}
        \vspace{2mm}
        
        \begin{subfigure}{4.75in}
            \includegraphics[trim={0mm 0mm 0mm 0mm},clip, height=44mm, width=59mm]{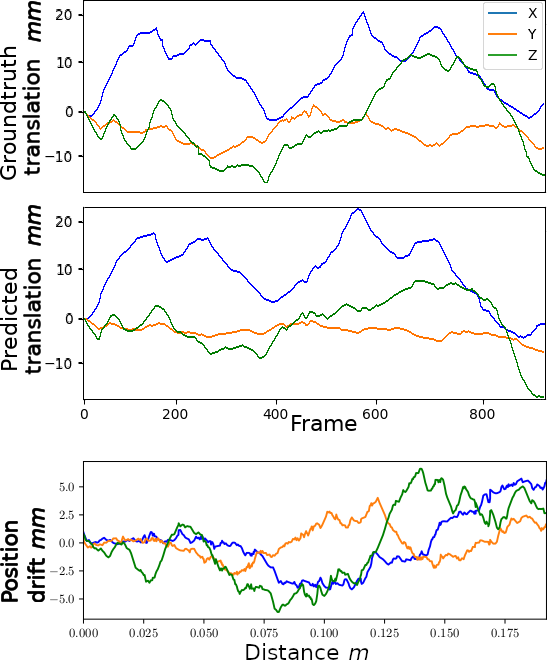} \hspace{0mm} \includegraphics[trim={0mm 0mm 0mm 0mm},clip, height=44mm, width=59mm]{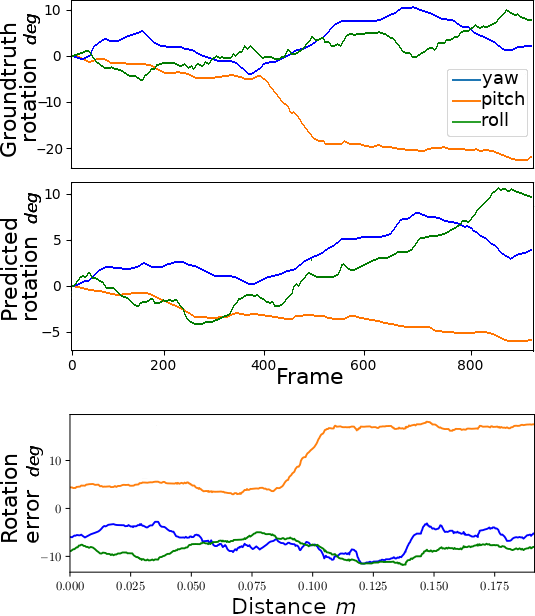} 
            \subcaption{Qualitative result}
        \end{subfigure}    
        \caption{The validation losses are shown in subfigure (a) for translation, (b) for the angle, and (c) for photometric reprojection loss. The qualitative results are showing the actual camera poses for sheep joint in (d). The first row of (d) is the groundtruth, the middle row is the corresponding network predictions (supervised+self-supervised stereo) and the third row is the ATE. The left column is the translation and the right column is the angle (rotation). }
\label{Plots} 
\end{figure}

\vspace{-2mm}
The training and validation losses of the depth+pose networks are shown in Fig. \ref{Plots}. For comparison between the supervised+self-supervised depth+pose estimation using images at time t+1 and the stereo pair, we included the plot for the self-supervised counterpart as well as supervised+self-supervised using image at time t+1 only, i.e., mono supervised+self-supervised. In this way, it is possible to evaluate the impact of the pose supervision and stereo versus mono scenario.

According to the plots on the training and validation data, Fig. \ref{Plots}(a) and (b) respectively, the photometric reprojection loss (which is an indication of the combined accuracy of camera pose and the depth estimation) is lowest for the self-supervised network. It is closely followed by the supervised+self-supervised stereo networks. The supervised+self-supervised mono has the poorest outcome with the photometric reprojection loss. On the other hand, the supervised+self-supervised networks outperformed the self-supervised network on the camera pose estimation for the validation data as shown in Fig. \ref{Plots}(c) and (d). 
These results from Fig. \ref{Plots} indicate that the supervised+self-supervised networks using the stereo pairs and time t+1 as reference images have higher accuracy in depth and slightly higher on camera pose estimation compared with the supervised+self-supervised mono. Hence, this model was considered for 3D mapping of the scene.

\begin{figure}
        \begin{subfigure}{2.50in}
            \includegraphics[trim={0mm 0mm 0mm 0mm},clip, scale=0.28]{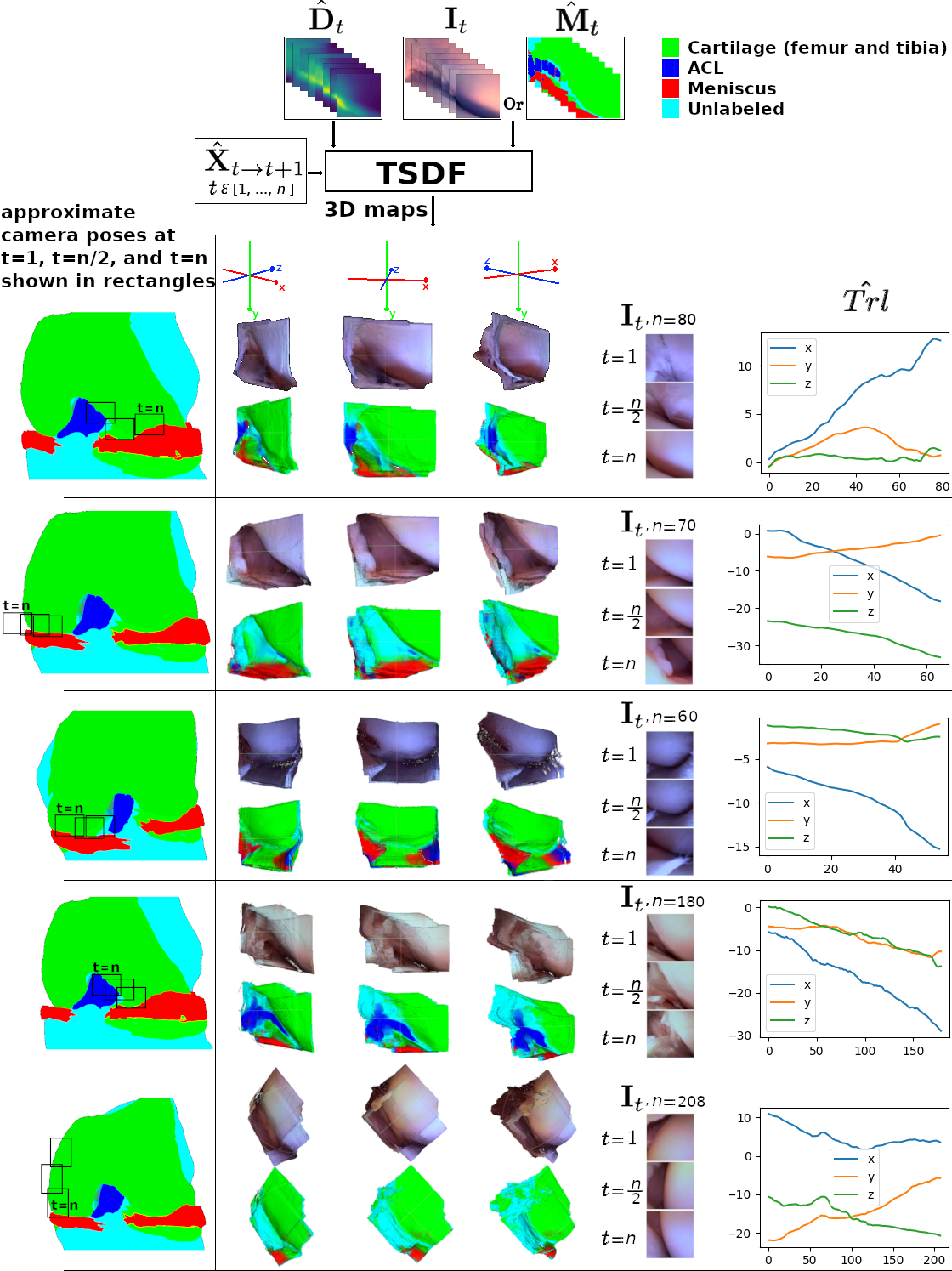} \hspace{0mm}
        \end{subfigure} %
        \vspace{2mm}
        \caption{Sample 3D maps on the test data from actual human knee. Once the networks are trained, the arthroscopic frames are provided as input to them and their output of depth $\hat{\mathbf{D}}_t$ and camera pose, $\hat{\mathbf{X}}_{t\rightarrow t+1}$, will be given as inputs to the TSDF function~\cite{zach_globally_2007} to create the extended map. Either the image $\hat{\mathbf{D}}_t$ or the corresponding semantic label, $\hat{\mathbf{M}}_t$, can be provided as the 3rd input to the TSDF function. The variable $n$ refers to the number of the frames used to create the corresponding map. Since most of the variation in $\hat{\mathbf{X}}$ is due to translation, only the $\hat{\mathbf{Trl}}$ was shown in the figure. The knee models on the left hand side of the figure with dark squares, show the approximate locations of the camera with respect to the actual knee.  }  
\label{Maps} 
\end{figure}

The actual pose network predictions on the validation data (animal experiment) are shown in Fig. \ref{Plots}(d). The groundtruth is shown in the first row while the prediction is on the second row and the corresponding ATE on the 3rd row.
In general, the changes in rotation proved to be harder for the network to predict than translation. Overall, the prediction of the camera rotation was more difficult than the translation. 
Fig. \ref{Maps} shows sample 3D maps obtained by fusing chunks of arthroscope frames from a human cadaver knee. The number of frames to create the maps is indicated by $n$. The corresponding camera translations are shown in the plots on the right side of each map. For every map, the semantic map is also provided with green being the cartilage (femur and tibia), meniscus in red, and ACL in blue. The cyan refers to other structures such as typically floating fat and skin. The segmentations appear correct except for a minor error in the third map where the segmentation network falsely detects meniscus (red) on the right hand side of the map.

\vspace{-2mm}
\section{Conclusion}
\vspace{-2mm}
In this work for the first time, we presented a pipeline to perform 3D semantic mapping in arthroscopy. To the best of our knowledge, this has not been done in any medical endoscopy before. To achieve these, we used the deep learning approaches for semantic segmentation, depth perception and camera pose estimation. 
The proposed domain adaptive approach produced superior accuracy in camera pose estimation and comparable depth accuracy in comparison with the self-supervised counterpart. 
Furthermore, we used a segmentation tool to semantically segment the images into cartilage, ACL, and meniscus. The segmentation approach utilizes the multi-spectral properties of surgical tissue in the images rather than merely the geometrical cues as was shown in~\cite{jonmohamadi_automatic_2020,liu_self-supervised_2020}.
\section*{Acknowledgements}
Supported  by  AISRF53820 and Australian  Research  Council  through  grants  DP180103232  and FT190100525.
\bibliographystyle{splncs04}
\bibliography{Ref_3DMap}

\end{document}